 \documentclass[pmlr,twocolumn,10pt]{jmlr} 

\usepackage{booktabs}
\usepackage[load-configurations=version-1]{siunitx} 

\theorembodyfont{\upshape}
\theoremheaderfont{\scshape}
\theorempostheader{:}
\theoremsep{\newline}

\def \MIM{{\hyperref[MIMDef]{MIM}}}

\jmlrvolume{LEAVE UNSET}
\jmlryear{2021}
\jmlrsubmitted{LEAVE UNSET}
\jmlrpublished{LEAVE UNSET}
\jmlrworkshop{Machine Learning for Health (ML4H) 2021} 

\title[MIM]{Multi network InfoMax: A pre-training method involving graph convolutional networks}

  \author{%
   \Name{Usman Mahmood} \Email{umahmood1@student.gsu.edu}\\
   \Name{Zening Fu} \Email{zfu@gsu.edu}\\
   \Name{Vince Calhoun} \Email{vcalhoun.gsu.edu}\\
   \Name{Sergey Plis} \Email{s.m.plis@gmail.com}\\
   \addr Tri-institutional Center for Translational Research in Neuroimaging and Data Science (TReNDS), Georgia State University, Georgia Tech, Emory
  }
  
\begin{document}

\maketitle

\begin{abstract}
Discovering distinct features and their relations from data can help us uncover valuable knowledge crucial for various tasks, e.g., classification. In neuroimaging, these features could help to understand, classify, and possibly prevent brain disorders. Model introspection of highly performant overparameterized deep learning (DL) models could help find these features and relations. However, to achieve high-performance level DL models require numerous labeled training samples ($n$) rarely available in many fields. This paper presents a pre-training method involving graph convolutional/neural networks (GCNs/GNNs), based on maximizing mutual information between two high-level embeddings of an input sample. Many of the recently proposed pre-training methods pre-train one of many possible networks of an architecture. Since almost every DL model is an ensemble of multiple networks, we take our high-level embeddings from two different networks of a model --a convolutional and a graph network--. The learned high-level graph latent representations help increase performance for downstream graph classification tasks and bypass the need for a high number of labeled data samples. We apply our method to a neuroimaging dataset for classifying subjects into healthy control (HC) and schizophrenia (SZ) groups. Our experiments show that the pre-trained model significantly outperforms the non-pre-trained model and requires $50\%$ less data for similar performance.
\end{abstract}

\section{Introduction}
\label{sec:intro}

Classical machine learning algorithms (support vector machine (SVM), logistic regression (LR), etc.) have proven to be highly efficient in classification tasks \cite{article,5508526}. These algorithms work on hand-crafted features, e.g., functional network connectivity (FNC)  matrices~\cite{yan2017discriminating}, and produce state-of-the-art results~\cite{DOUGLAS2011544}. Modern biomedical imaging, functional MRI, collects high dimensional data, where the number of measured values ($m$) per sample can exceed tens of thousands.  Machine learning algorithms does not provide good classification performance in this case as the data dimensions are much higher than the number of data samples ($n$) available for training, thus creating the curse of dimensionality ($m \gg n$). Methods of acquiring hand-crafted features, such as FNCs, work to reduce these dimensions. These hand-crafted features are then necessarily void of many valuable properties and dynamics initially present in the data. On the other hand, using data dynamics is essential for finding distinct features and their relations, responsible for classification. These dynamics are crucial for the neuroimaging field, where causes and functional connectivity between brain regions of the underlying brain disorders are still unclear. Finding the causes of disorders and the underlying brain regions connectivity can potentially help prevent, delay and even cure these disorders.

Introspection of highly performant deep learning (ML) models  could help in finding distinct features and their relations but this approach requires a significantly high number of training samples to work. With data dimensions $m$ exceeding thousands for fMRIs, it is very difficult for a single study to have enough data subjects $n$ to not have the problem of $m\gg n$.  To overcome this difficulty, researchers in many fields~\cite{henaff2019data,devlin2018bert,lugosch2019speech} traditionally employ un-supervised/self-supervised pre-training~\cite{erhan2010does}. Furthermore, self-supervised methods with mutual information objective are able to preform competitively  with supervised methods~\cite{infonce,hjelm2018learning,bachman2019learning} and are suitable for a number of applications \cite{anand2019unsupervised,ravanelli2018learning}. Pre-training and transfer learning have also been used for neuro/brain imaging  ~\cite{mensch2017learning,thomas2019deep,2020,10.3389/fnins.2018.00491}.

In this study, we propose a novel pre-training algorithm involving GNNs which we call \phantomsection\label{MIMDef} Multinetwork InfoMax (MIM). MIM uses self-supervised learning with a mutual information objective to learn data representations. These representations are then used for downstream task of classification using small labeled dataset. We show that using pre-training we can bypass the need to acquire large labelled training datasets and achieve higher classification performance than non pre-trained counter parts. Prior work on self-supervised pre-training for GNNs \cite{bojchevski2018deep,davidson2018hyperspherical,48921} only pre-trains the encoder represented by  one of the many DL architectures available today. However, many newly introduced models are an ensemble of multiple networks which work together for a single task and cannot be easily or sensibly split into an encoder and “the rest”. Unlike these existing approaches, our technique maximises the mutual information between two global embeddings, acquired from different parts of the architecture, with positive and negative samples coming from the input data. Thus, we pre-train the complete model rather than splitting it into constituent parts. We use the model presented in~\cite{2021}.

\section{MIM}
\label{Method}
We present \MIM{} which first learns representations using an unrelated and unlabeled dataset and then transfers those representations for downstream training and classification. The model does that by increasing relation and mutual information between the global embedding of $y$ and $c$ belonging to the same subject, acquired from the convolutional neural network (CNN) and GNN part of the model. While doing this, the CNN and GNN module get synced with each other. The model learns and creates representations of the subjects, which proves helpful for downstream classification by forcing the GNN output to perceive information from a complete graph.  

To accomplish this, we train a critic function that takes the two vectors $y$ and $c$, representing the global embedding of the subject, and assigns a higher value for positive pairs compared to negative pairs. In our experiments, \textit{positive pairs} are defined as embedding coming from the same subject, where \textit{negative pairs} are when $y$ and $c$ are from different subjects. More specifically, let $D = \{(y^i, {c}^j): 1 \le i,j \le N \}$ be a dataset of pairs. $y^i$ is the global embedding of the subject $i$, ${c}^j$ is the global embedding for the subject $j$. $N$ is the total number of subjects. Then $D^+ = \{(y^i, {c}^j ): i = j \}$ is called a dataset of positive pairs and $D^- = \{(c^i, {y}^j ): i \ne j \}$ a set of negative pairs. Using our critic function $f$ we define a lowerbound of InfoNCE estimator~\cite{infonce} $\mathcal{I}_f(D^+)$:

\begin{equation}
 \mathcal{I}(D^+) \ge \mathcal{I}_f(D^+) \triangleq \sum^N_{i=1}  \log \frac{\exp f((c^i, {y}^i))}{\sum^N_{j=1} \exp f((c^i, {y}^j))},
\end{equation}
where $f(c,y) = (c)(y)^\intercal$ and loss $L$ is InfoNCE; $L = - \mathcal{I}_{f}$. Figure \ref{fig:arch} shows the complete architecture.

\section{Experiments}
We show the advantages of pre-training the model using a real neuro-imaging dataset. We classify schizophrenia (SZ) patients from health controls (HC) with varying training sizes. We compare our model with \cite{2020,2021} and classical machine learning algorithms, SVM and LR.

\begin{figure}[t]
\floatconts
  {fig:arch}
  {\caption{The complete architecture \MIM{}. Note that some parts and links are just for pre-training.}}
  {\includegraphics[width=.8\linewidth]{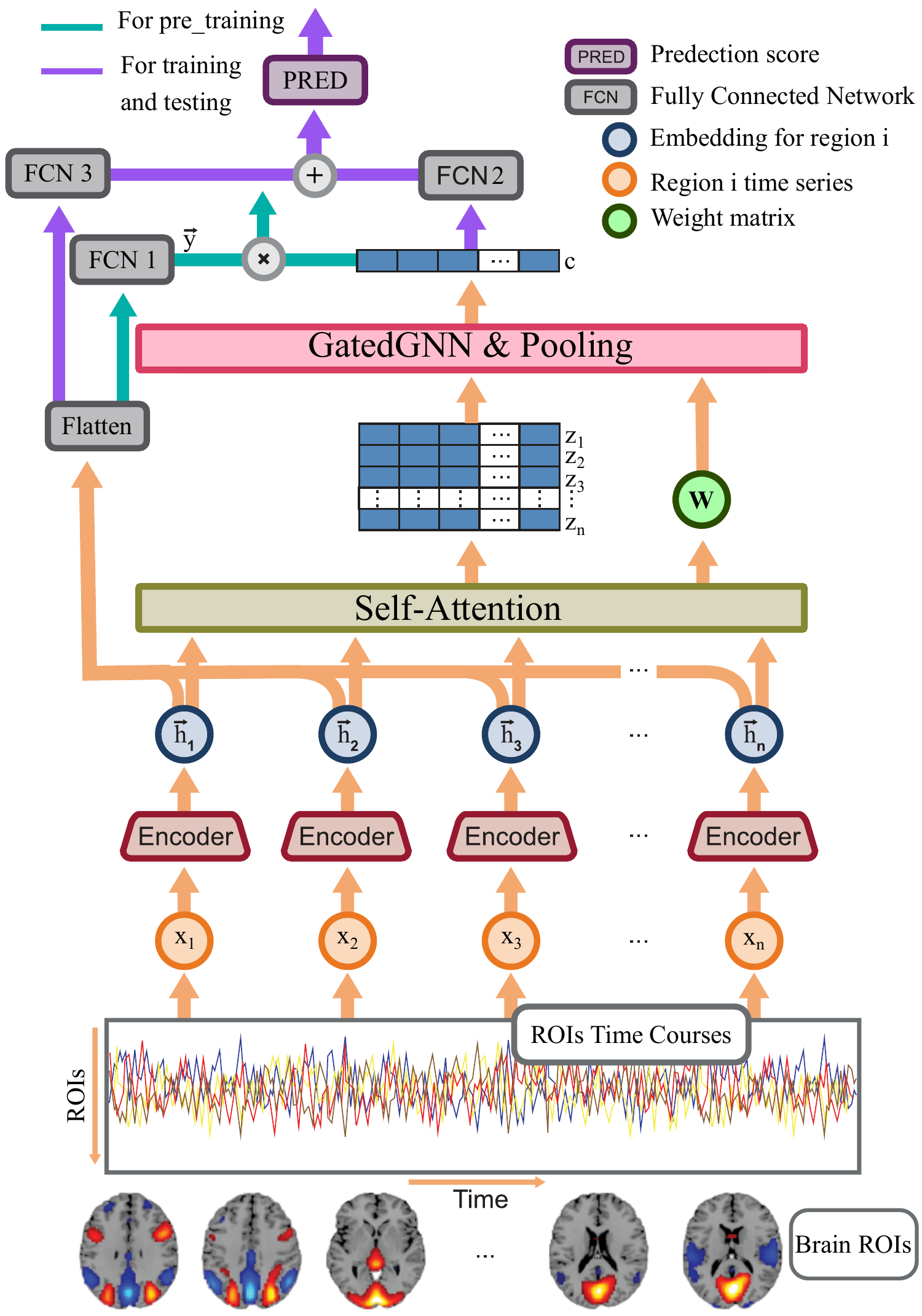}}
\end{figure}


\subsection{Dataset}
We worked with two datasets in our experiments. Human Connectome Project (HCP) ~\cite{van2013wu}, which has only HC subjects is used for pre-training, and the data from Function Biomedical Informatics Research Network (FBIRN)~\cite{keator2016function} dataset including both SZ and HC subjects is used for downstream training and testing our model. Resting fMRI data from the phase III FBIRN study were analyzed for this project. The dataset has $368$ total subjects.

\subsection{Pre\_processing}

We preprocess the fMRI data using statistical parametric mapping (SPM12, http://www.fil.ion.ucl.ac.uk/spm/) under MATLAB 2019 environment.
After the preprocessing, subjects were included in the analysis if the
subjects have head motion $\le 3^\circ$ and $\le 3$ mm, and with functional data providing near full brain successful normalization~\cite{fu2019altered}. This preprocessing results in $311$ subjects. The data is then divided into brain regions using the automated anatomical labeling (AAL) atlas \cite{TZOURIOMAZOYER2002273} and weighted average of voxels was computed inside each region, similar to \cite{2021}.

\subsection{Pre-training, Training and Testing}
We pre-train our model on a large unrelated and unlabeled dataset (HCP). After pre-training, we transfer the learned representations and further train the model to classify SZ and HC subjects using the fBIRN dataset. For downstream training and testing, we add a small linear layer on top for classification and train the complete architecture. The following section explains the architectural details of our model.

\subsubsection{Architecture details}
Each ROI time series vector of size $160$, is passed through a 1DCNN \cite{KIRANYAZ2021107398} to get an embedding of size $64$. The encoder consists of $4$ convolution layers with filter size $(4,4,3,1)$, and ReLU activation in between, stride $(2,1,2,1)$ and output channels $(32,64,64,10)$, and a fully-connected layer on top to get a vector of size $64$, denoted as $h_i$. 
The representations ($h_i$) of a subject are then passed through a self-attention \cite{10.5555/3295222.3295349} network. ($h_is$) of a subject are also flattened to get a full embedding of the subject, represented by $h_f$. During pre-training, $h_f$ is passed through a fully convolutional network (FCN) with $3$ layers of size ($1024$, $128$, $96$), to match the dimensions of $c$. The FCN gives a single vector (y) representing the global embedding of the subject created using the encoder. During downstream classification, $h_f$, is passed through FCN to get encoder-based predictions.


The self-attention network gives a new embedding of size $24$ for each region and the weights among regions. These regions and weights are treated as nodes and edge weights, thus giving a graph. 
The graph is then passed to gated-GNN \cite{li2016gated} with $6$ layers. The first $3$ layers are followed by topk-pooling \cite{gao2019graph,knyazev2019understanding} selecting top (80,80,30) percent nodes, respectively. Three global pooling layers follow the final graph layer; 1) global average pool, 2) global max pool, and 3) attention-based pooling \cite{vinyals2016order}. The pooling layers give us vectors of size $24$, $24$, and $48$, respectively. We concatenate the three vectors to create a single vector ($c$) of size $96$ representing the subject.

During pre-training, we increase the mutual information between two global embeddings (y) and (c) as explained in section \ref{Method}. During downstream training and testing, pre-training part is skipped, and (c) is passed through an FCN with $2$ layers of size $32$ and $2$ for classification summing it with the FCN output of $h_f$ obtained by the encoder. This is done to train the encoder to give classification-based representation by directly applying the gradients during back-propagation. Orthogonal and Xavier normal (gain=$0.25$) initialization \cite{GlorotAISTATS2010} was used for encoder and rest of the model.

\subsection{Results}

We perform different experiments with our model to show how pre-training can increase classification performance in the downstream task. In all the experiments, we compute the AUC (Area Under Curve) to measure performance. For all experiments we perform 18 fold cross validation with validation and test size of $17$ subjects. We conduct $10$ randomly seeded trials for each experiment. Early stopping was used based on validation data with a learning rate of $1e-4$.

\subsubsection{Varying training size}

Figure \ref{fig:AUC UFPT and NPT} show the difference in classification performance of our pre-trained model \MIM{} with non-pre-trained model BrainGNN \cite{2021}. The pre-trained model (\MIM{}) performs much better than BrainGNN with a smaller training size and almost reaches the maximum score of BrainGNN with $50\%$ fewer training subjects.

\subsubsection{Comparison with other methods}

We compare \MIM{} with two popular machine learning models. Figure \ref{fig:AUC All Comp 19 fold} shows the results of our experiments. It is clear that traditional ML methods does not perform well directly on rs-fMRI data because of $m>>n$ problem. \MIM{} performs better than wholeMILC, which is another pre-trained method. ML methods with handcrafted features FNCs as input perform well and result in $\sim 0.81$ AUC. FNCs are computed using Pearson correlation coefficient (PCC). Machine learning models' results are computed using python package Polyssifier, available at \url{https://github.com/alvarouc/polyssifier}.

\begin{figure}[t]
\floatconts
  {fig:AUC UFPT and NPT}
  {\caption{\MIM{} gives almost the same results as BrainGNN with half of the training subjects. The stars show the significance of the difference between the classification result of the two models. Significance is the measure of p-value of t-test.)}}
  {\includegraphics[width=\linewidth]{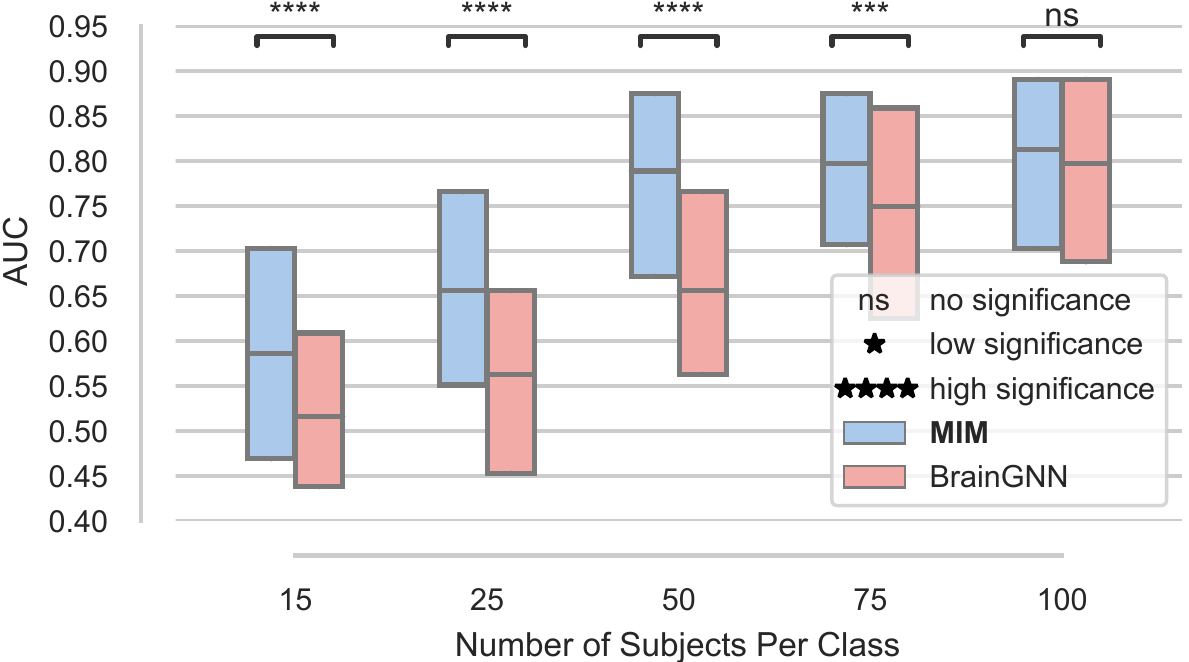}}
\end{figure}

\begin{figure}[ht]
\floatconts
  {fig:AUC All Comp 19 fold}
  {\caption{Comparison with DL and ML algorithms show that \MIM{} is the best performing model in average AUC score using $100$ subjects per class for training. The machine learning models completely fail on region data which was used for the deep learning models(\MIM{}, BrainGNN, WholeMILC).}}
  {\includegraphics[width=\linewidth]{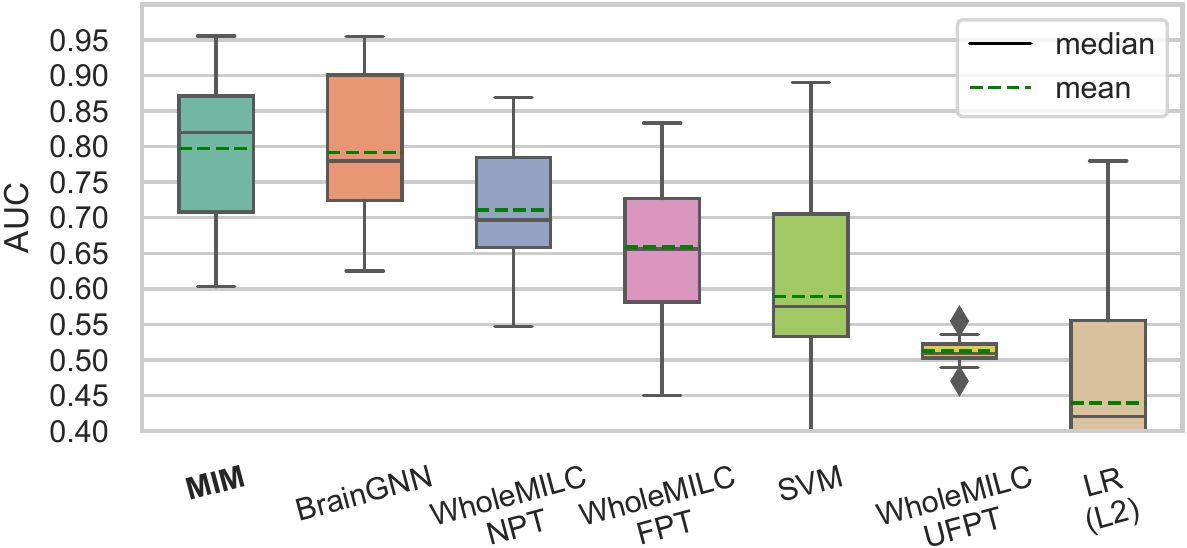}}
\end{figure}




\section{Conclusion}
We show that pre-training the complete architecture, composed of different networks (CNN, GNN e.t.c.) on an unlabeled dataset can significantly improve the classification performance on a neuroimaging dataset. Pre-training the model in a self-supervised way and then using transfer learning for the downstream task can help mitigate the problem of small labeled datasets. Pre-training helps the model to learn data dynamics which results in acquiring more generalized representations. The benefits would be more profound when applied directly to spatio-temporal data (fMRI). Region-based data used in this study had fewer data dimensions ($m$) than fMRI. Dataset with fewer dimensions allows the non-pre-trained model to perform well but is void of many essential data dynamics. Using pre-training directly on the fMRI dataset can eliminate the need for any dimension reduction method and the benifits would be profound. 

\acks{Data for Schizophrenia classification was used in this study were acquired from the Function
BIRN Data Repository (http://bdr.birncommunity.org:8080/BDR/),
supported by grants to the Function BIRN (U24-RR021992) Testbed funded by the
National Center for Research Resources at the National Institutes of Health, U.S.A. and from the Collaborative Informatics and Neuroimaging
Suite Data Exchange tool (COINS; \href{http://coins.trendscenter.org}{http://coins.trendscenter.org}). This work was in part supported by NIH grants 1RF1MH12188 and 2R01EB006841.}

\bibliography{jmlr-sample}

\appendix





\end{document}